\documentclass[runningheads]{llncs}
\usepackage{hyperref}
\usepackage{graphicx}
\usepackage{multirow}
\usepackage{amsmath}
\usepackage{xcolor}
\usepackage{booktabs}
\usepackage{caption}
\usepackage{subcaption}
\usepackage{amsfonts}
\usepackage{fixltx2e}
\usepackage{tabularx}
\usepackage[ruled]{algorithm2e}
\usepackage{xcolor}
\usepackage{graphicx}
\usepackage{latexsym}
\SetKwComment{Comment}{/* }{ */}

\begin{document}
\title{Generating Explanations in Medical Question-Answering by Expectation Maximization Inference over Evidence}
%
%
\author{Wei Sun\inst{1} \and
Mingxiao Li\inst{1} \and
Damien Sileo\inst{2} \and
Jesse Davis\inst{1} \and
Marie-Francine Moens\inst{1}
}%
\authorrunning{F. Author et al.}
%
\institute{KU Leuven, Leuven, Belgium \\
\email{firstname.lastname@kuleuven.be}
\and  Inria Centre, Lille \\
\email{damien.sileo@inria.fr}}
\maketitle              
\begin{abstract}
Medical Question Answering~(medical QA) systems play an essential role in assisting healthcare workers in finding answers to their questions.  
However, it is not sufficient to merely provide answers by medical QA systems because users might want explanations, that is, more analytic statements in natural language that describe the elements and context that support the answer. 
To do so, we propose a novel approach for generating natural language explanations for answers predicted by medical QA systems.
As high-quality medical explanations require additional medical knowledge, so that our system extract knowledge from medical textbooks to enhance the quality of explanations during the explanation generation process.
Concretely, we designed an expectation-maximization approach that makes inferences about the evidence found in these texts, offering an efficient way to focus attention on lengthy evidence passages.
Experimental results, conducted on two datasets MQAE-diag and MQAE, demonstrate the effectiveness of our framework for reasoning with textual evidence. 
Our approach outperforms state-of-the-art models, achieving a significant improvement of \textbf{6.86} and \textbf{9.43} percentage points on the Rouge-1 score; \textbf{8.23} and \textbf{7.82} percentage points on the Bleu-4 score on the respective datasets.

\end{abstract}
\section{Introduction}

Medical Question Answering (Medical QA) is an essential component of modern healthcare. By automating the process of answering medical queries, it spares users the effort of manual searches. 
Despite the development of advanced medical QA systems~\cite{abacha2015means,veisi2020persian}, the ideal medical QA systems should not only provide accurate answers but also explanations that describe the elements and context that support a choice~\cite{jin2022biomedical}.
Explanations serve as a linchpin for promoting the reliability of answers and facilitating fact-checking~\cite{jin2022biomedical,kotonya-toni-2020-explainable-automated}, so that users, i.e., patients and clinicians, can trust medical QA systems by addressing their concerns.
An alternative explanation generation approach is to generate concise, self-contained natural language explanations that are more comprehensible and accessible to users~\cite{chen2022learning}.
This approach can empower clinicians and patients to better understand the medical QA system's behavior, facilitating informed decisions and improved patient care~\cite{lee2006beyond}.

In this paper, we will explore what we call \textbf{Q}uestion-\textbf{A}nswering \textbf{E}xplanation~(\textbf{QAE}), where the goal is to provide a natural language explanation of the answer or decision in a medical QA system.  
QAE provides explanations for the returned answer given a question, a patient's case description, and possibly external data.
Prior works~\cite{know_bai,hua-wang-2018-neural} demonstrate that evidence retrieved from an external knowledge base helps to improve the quality and effectiveness of NLP systems.
For medical explanation generation, a key challenge to improve the quality of explanations lies in how to fuse medical knowledge from multiple lengthy documents This is difficult because it is often unclear which evidence pieces are relevant and each document is quite long. 
Classical multi-head attention mechanisms struggle to deal with multiple lengthy documents.
It is expensive to conduct full attention across concatenated evidence because their time complexity increases exponentially as the number of input tokens increases. 
To overcome this challenge, we propose \textbf {E}xpectation-\textbf{M}aximization \textbf{I}nference over evide\textbf{N}ces~(\textbf{EMIN}), implemented as an encoder-decoder model with a dual encoder for encoding question and answer, and evidence texts.  
This approach iteratively optimizes both the relevance weights of each piece of evidence and the text generation model's parameters. This permits attending over separated evidence rather than performing full attention across concatenated evidence.

Our main contributions are the following: 
\begin{itemize}
    \item We introduce a novel task known as Question-Answering Explanation~(QAE) in medical domain.
    We present two new datasets, MQAE-diag and MQAE, curated from publicly available medical resources. 
    \item We propose a novel framework, Expectation-Maximization Inference over evidences (EMIN), designed for generating contextualized explanations.
    EMIN enables inference of the relevance of each piece of evidence and leverages attention in a scalable way. 
    \item Experimental results show that our model outperforms state-of-the-art baselines by a large margin on the MQAE and MQAE-diag datasets across different evaluation scores.
    Notably, our model achieved improvements by \textbf{6.86} and \textbf{9.43} percentage points on the Rouge-1 score; \textbf{8.23} and \textbf{7.82} percentage points on the Bleu-4 score on the MQAE-diag and MQAE datasets. 
\end{itemize}

This paper is organized as follows: 
Section~\ref{sec:methods} details the EMIN model.
Section~\ref{sec:exp} we presents results on two medical QAE datasets and an error analysis.
Finally, section~\ref{sec:related} describes related works and section~\ref{sec:conclude} summarizes this paper.

\begin{figure*}[htbp]
\centering
\includegraphics[width=\linewidth]{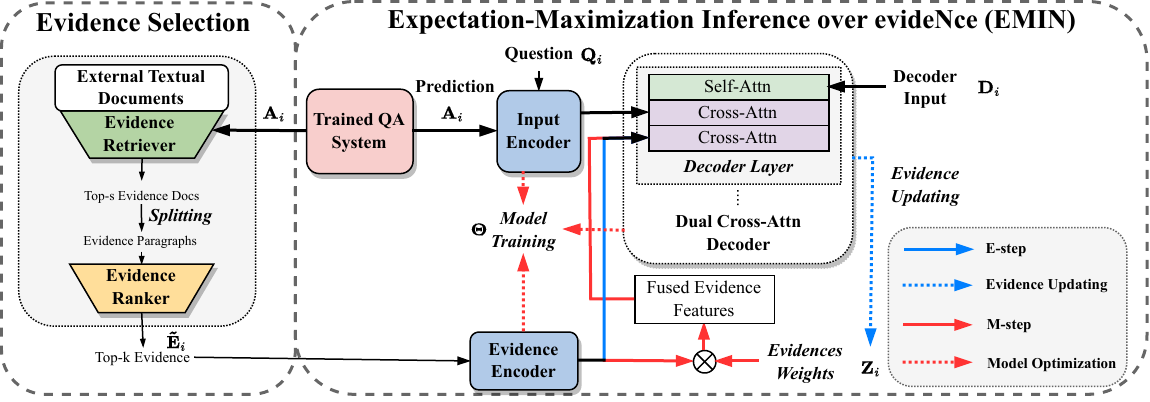}
\caption{The overall architecture of the EMIN model~(training stage). 
The evidence query is the question~$Q_i$'s answer. 
Model parameters consist of the parameters of the question, evidence encoders and dual cross-attention decoder. 
We 
use of ground-truth answers during the training phase, while relying on the predictions generated by a trained QA system during the inference phase. }
\captionsetup{justification=centering}
\label{fig:architecture}
\end{figure*}

\section{Methods}
\label{sec:methods}

The goal of \textbf{Q}uestion-\textbf{A}nswering \textbf{E}xplanation~(\textbf{QAE}) is to generate a natural language explanation of the predicted answer to a given question.
We utilize the ground-truth answer for training purposes and rely on the predictions generated by a trained QA system during the inference phase. 
We will approach this as a text-to-text generation task.  
More formally, for training we assume access to a dataset {((${Q}_1$, ${A}_1$, $E_1$),${Exp}_1$),$\cdots$,((${Q}_N$, ${A}_N$, $E_N$),${Exp}_N$)}, where ${Q}_i$ is a question,  ${A}_i$ is the answer to this question, 
$E_i$ refers to the set of relevant evidence paragraphs and ${Exp}_i$ is the corresponding explanation.  
The goal is to learn a question-answering system $\mathbf{f}^\prime: {Q}_i \rightarrow {A}_i$ and an explanation generator $\mathbf{f}: \{{Q}_i, {A}_i, {E}_i\} \rightarrow {Exp}_i$ where the $Exp_i$ is a text composed of natural language sentences.

Figure~\ref{fig:architecture} shows the overall architecture of our approach, which works as follows. 
First, we automatically identify a limited set of potential pieces of evidence in a document collection, typically represented as relevant paragraphs corresponding to each question.
Second, we employ a text generation model based on a BART encoding-decoding neural architecture~\cite{lewis2019bart}. 
We fine-tune this model using an expectation-maximization (EM) approach. This iterative process refines not only the model's parameters, including attention parameters but also updates the importance weights associated with each potential evidence piece.
During inference, we employ an analogous EM-based approach to generate explanations for question-answer pairs that have not been previously encountered. 
In this phase, we rely on a trained question-answering system to predict the answers.
The following subsections describe each component of the approach in detail.


\subsection{Evidence Selection}
\label{sec:evidence_select}
The initial step involves selecting a set of evidence paragraphs from a document collection that are likely to be relevant for generating the explanation.
To do so, we employ the Dense Passage Retrieval~(DPR) model~\cite{karpukhin2020dense}, which retrieves documents related to the answer from a medical document collection (in our case medical textbooks).
The retrieval query is the answer ${A}_i$.
Before initiating the retrieval process, both documents and retrieval queries undergo preprocessing. 
This involves tokenization into words, removal of stop words, and conversion of all remaining words to lowercase. 
The DPR algorithm subsequently returns a set of evidence documents based on their relevance scores.
Following the retrieval of documents, the next step involves breaking them down into evidence paragraphs. 
These evidence paragraphs are then ranked using the DICE similarity score concerning the textual retrieval query.
The top-$k$ ranked textual evidence paragraphs are selected, taking into account normalized word overlap.\footnote{This approach resulted in better explanation generation results than first splitting the documents into evidence paragraphs and then ranking them with DPR.} 

\subsection{Neural Backbone}
\label{sec:nn}
The dual encoder-decoder model is based on BART~\cite{lewis2019bart}. It has two encoders: one encodes the question-answer and the other encodes the evidence paragraphs. 
We initialize the dual encoder-decoder model using the pretrained parameters of the BART base model.

\subsubsection{Encoder}
\label{sec:nn-enc}
The BART encoder has a transformer architecture characterized by applying self-attention to the representations at a specific layer $l$ of the transformer encoder.
Let $X_i$ be the concatenation of the question string~$Q_i$ and answer string~$A_i$. 
We pass $X_i$ to the encoder and use $\mathbf{X}_i^l \in \mathbb{R}^{L_x \times d}$ to represent its output at a specific layer $l$ where
$L_x$ is the sequence length of~$X_i$ and $d$ is the dimension of the hidden states.
Evidence features~$ {\mathbf{E}}_i 
$ are obtained by feeding each selected evidence paragraph $\mathbf{E}_{i,j} \in \mathbb{R}^{L_e \times d}$ into the evidence encoder. 
${\mathbf{E}}_i^l$ refers to the encoded evidence paragraphs at a specific layer $l$ of the transformer encoder.
$L_e$ represents the evidence's sequence length.



\subsubsection{Dual Cross Attention Decoder}
\label{sec:dual_cross}
\paragraph{Vanilla BART decoder} 
The BART decoder has a transformer architecture characterized by self-attention and cross-attention over the tokens of the encoder. The decoder is composed of different layers. 
Given the hidden state of the $l$th layer, $\mathbf{H}^{l} \in \mathbb{R}^{L_d \times d}$, which represents the hidden states of the decoder sequence.
$L_d$ is the length of the decoder sequence in terms of tokens.
The hidden state of the next layer $\mathbf{H}^{l+1}$ is computed as:
\begin{align}
\label{eq:E_decoder_layer}
\mathbf{SA_i^{l}} &= \operatorname{Self-Attn} \Bigl ( \mathbf{H}^{l}_i, \mathbf{H}^{l}_i \Bigl ),\\
\mathbf{CA}_i^{l} &= \operatorname{Cross-Attn} \Bigl ( \mathbf{SA}_i^{l}, \mathbf{X}_i^l \Bigl ),
\end{align}
where $\operatorname{Self-Attn}$ and $\operatorname{Cross-Attn}$ operations compute the scaled dot-product attention~\cite{vaswani2017attention}.
The decoder autoregressively generates explanations. 
That is, at each timestep of the decoding process, the input of the decoder is the embedding of the previous token (embedding of the ground truth token during training and embedding of the generated token during inference).

\paragraph{EMIN decoder} 
In EMIN, we explicitly model evidence weights as $\mathbf{z}_{i} \in \mathbb{R}^{k}$ that capture the relevance of each selected evidence paragraph for instance $i$. As we do not have ground truth evidence weights, we infer them using an expectation-maximization algorithm. The inferred evidence weights make the attention of many texts or on a long text split in paragraphs scalable for realistic applications. 

The EMIN decoder follows the general architecture of the vanilla BART decoder. 
This entails that the hidden state of each layer is updated with the context vector that contains the encoded information of the evidence tokens weighted by the cross-attention weights obtained by $\operatorname{Cross-Attn} \Bigl (\mathbf{CA}_i^{l}, \mathbf{E}_i^l \Bigl )$. 
In EMIN, each evidence paragraph $\mathbf{E}_{i,j}$ is also weighted with its evidence weight $\mathbf{z}_{i,j}$ by element-wise multiplication of their respective vectors. 

The intuition is that every transformer layer of the decoder cross-attends to relevant feature extracted by the corresponding encoder layer builds a representation to be decoded that captures different linguistic information. 
The representation of each layer can be used to update the evidence weight $\mathbf{z}_{i,j}$. 
Equation~($3$) becomes: 
\begin{align}
\label{eq:E_decoder_layer_1}
\mathbf{H}_i^{l+1} &= \mathbf{z}_{i} \odot \operatorname{Cross-Attn} \Bigl ( \mathbf{CA}_i^{l}, \mathbf{E}_i^l \Bigl ).
\end{align}

\subsection{Expectation-Maximization Inference over Evidence}
\label{sec:emin}
The evidence weights~$\mathbf{z}_i \in \mathbb{R}^{ k}$ capture the importance of the evidence paragraphs ~${\mathbf{E}}_i$. 
The evidence weights are the hidden variables in the EM algorithm with $\operatorname{sum}(\mathbf{z}_i) = 1$.
During training for each question-answer input $X_i$, the weights of the evidence paragraphs $\mathbf{z}_{i}$ and parameters $\mathbf{\Theta}$ of the dual encoder-decoder model are iteratively optimized by the EM algorithm.  
In the E-step, we update the evidence weights $\mathbf{z}_{i}$ using the ground-truth explanations.
In the M-step, we use the back-propagation algorithm to optimize the dual encoder-decoder model's parameters $\mathbf{\Theta}$ given the expected values of $\mathbf{z}_{i}$.
During inference, we only iteratively update the evidence weights 
$\mathbf{z}_i$ based on the trained explanation generation model to gain insight into the contribution of each evidence paragraph in generating the explanation. 
The weights of the evidence paragraphs $\mathbf{z}_{i}$ of an input example are uniformly initialized as $\frac{1}{k}$ unless mentioned otherwise.

\subsubsection{Expectation-Maximization during Training}
\label{sec:emi_train}

\noindent \\ \textbf{\textit{E-step:}} 
In the E-step at each time step $t$, given input features~$\mathbf{X}_i$, the parameters of the dual encoder-decoder model at time step $t-1$,
and evidence representation~$\mathbf{{E}}_i=[\mathbf{{E}}_{i,1},\mathbf{{E}}_{i,2},\dots,\mathbf{{E}}_{i,k}]$, we 
generate explanation $\overline{\mathbf{Exp}}_{i,j}^{t-1}$. 
We first compute the inverse cross-entropy between the generated and ground truth explanations:  
\begin{align}
 \label{eq:E_update_c}
    c_{i,j}^t = \frac{1}{-\sum_{m=1}^{L_d}P(Exp_{i,m}^{t-1})log(P(\overline{Exp}_{i,j,m}^{t-1}))},
\end{align}
\noindent where $P(\overline{Exp}_{i,j,m})$ is the generated token probability 
at position $m$ of sample $i$ when only evidence $j$ is used, and $P(Exp_{i,m})$ denotes the ground truth token 
at the same position of sample $i$, which is just a one-hot encoding in our case.

The expected values of $\mathbf{z}_i^t$ are then computed as: 
\begin{align}
\label{eq:E_update_z}
z_{i,j}^t &= \frac{\exp^{\frac{c_{i,j}^t}{\lambda}}}{\sum_{j=1}^{k} \exp^{\frac{c_{i,j}^t}{\lambda}}} , 
\end{align}
where $\lambda$ is a temperature parameter that controls the softness of the evidence weights distribution.
We gradually reduce $\lambda$ using $\lambda = e^{-0.01 \times \text{iteration}}$, so that weights will increasingly concentrate on the most important pieces of evidence. 

Intuitively, a piece of evidence should have a high weight if using it produces an generated explanation that is close to the ground truth one.
We leverage the inverse cross entropy to measure the similarity. The lower the value of this metric is, the closer the distance between the generated explanation and the ground-truth is, meaning that the model should pay more attention to this evidence paragraph by assigning it a larger weight. 

\noindent \textbf{\textit{M-step:}} In the M-step, at iteration step $t$ given the 
input features~$\mathbf{X}_i$, and the output ground truth explanation $Exp_i^t$, 
we optimize the dual encoder-decoder model's parameters~$\mathbf{\Theta}^{t}$ with backpropagation using the cross entropy~(CE) loss commonly used in text generation:
\begin{align}
\label{eq:M_update_theta}
\mathcal{L} &= - \sum_{i=1}^{N}\sum_{m=1}^{L_d}P(Exp_{i,m}^t)logP(\overline{Exp}_{i,m}^t),
\end{align}
where $L_d$ refers to the length of the ground truth sequence. 
$N$ is the number of training instances.
Note that during explanation generation, the decoder integrates the evidence weights computed in the E-step in the way described in section \ref{sec:dual_cross}.

The EM iteration stops when the 
KL divergence between the evidence weights of iteration $t$ and those of iteration $t-1$ is smaller than a defined threshold~$\epsilon$. 

\subsubsection{Inference}
\label{sec:dec}
When generating an explanation at test time, the model parameters $\mathbf{\Theta}^{t}$ are fixed, but we still need to estimate the weights of the evidence paragraphs $\bf{z}_i$ for the test question.
Here, we use the generated explanations to infer the weights of the evidence paragraphs. 
In each iteration $t$, we calculate the expected weight of the evidence in a manner similar to the \textbf{E-step} used during training. 
However, we modify this step by evaluating the cross-entropy of the generated explanation
and using the predicted answer from a trained question-answering system, namely BioLinkBERT~\cite{yasunaga2022linkbert}.
The global explanation generated at $t-1$: 

\begin{align}
\label{eq:E_update_c_infer}
c_{i,j}^t &= \frac{1}{-\sum_{m=1}^{L_d}P(\overline{Exp}_{i,m}^{t-1})logP(\overline{Exp}_{i,j,m}^{t-1})},
\end{align}


\begin{align}
\label{eq:E_update_z_infer}
z_{i,j}^t &= \frac{\exp^{\frac{c_{i,j}^t}{\lambda}}}{\sum_{j=1}^{k} \exp^{\frac{c_{i,j}^t}{\lambda}}}. 
\end{align}
The stopping criterion is the same as in the training phase. 
After $\bf{z}_i$ has converged, we generate the final explanation.

\section{Experiments}
\label{sec:exp}

\subsection{Datasets}
In our study, we utilize two medical question-answering explanation datasets, namely the Medical Diagnosis QAE (MQAE-diag) and the Medical QAE (MQAE) datasets, in conjunction with a repository of textual medical evidence documents. 
The distinction between MQAE-diag and MQAE lies in the nature of their questions. 
MQAE-diag is tailored for questions related to patient diagnosis, whereas MQAE is designed to evaluate fundamental medical knowledge. \footnote{Datasets and code will be made publicly available upon acceptance.}


\noindent \textbf{MQAE-diag} 
The MQAE-diag dataset's samples are derived from the MedMCQA dataset~\cite{pal2022medmcqa}, and two medical examination databases, i.e.,  the Spanish medical residency exam~(MIR)\footnote{\url{https://www.curso-mir.com/index.html}} and the United States Medical Licensing Examination (USMLE).\footnote{\url{https://www.usmle.org/}}
We select samples whose questions contain patients' descriptions and construct the MQAE-diag dataset.
In this dataset, there are 8038 samples for training, 995 instances for validation and 987 for testing. 
\\
\noindent \textbf{MQAE} 
The MQAE dataset is based on the MedMCQA dataset~\cite{pal2022medmcqa}. 
We extract 15993, 2000, and 1999 clinical instances for training, validation and testing.
In this dataset, the question only contains a single question and does not include any descriptions of the patient's case. \\
\noindent \textbf{Document Collection with Evidence Information. } We leverage the StatPearls medical book\footnote{\url{https://www.ncbi.nlm.nih.gov/books/NBK430685/}}, which contains 9070 medical documents, as our medical textual knowledge. 
Evidence retrieval is discussed in section \ref{sec:evidence_select}.

\begin{table*}[htbp]
\setlength\tabcolsep{4pt} 
\begin{center}
\begin{tabular*}{0.9\linewidth}{@{\extracolsep{\fill}} l r|r|r|r }
\toprule
Models & \multicolumn{1}{l}{R1} & \multicolumn{1}{l}{R2} & \multicolumn{1}{l}{RL} & \multicolumn{1}{l}{BLEU-4} \\
\midrule
\multicolumn{1}{l}{\textbf{T5}\textsubscript{q+a}}& 15.48$\pm$1.27 & 4.74$\pm$0.88 & 12.20$\pm$1.02 & 2.18$\pm$0.75  \\
\multicolumn{1}{l}{\textbf{BART}\textsubscript{q+a}} & 18.57$\pm$1.95 & 5.71$\pm$1.16 & 13.02$\pm$1.29 & 0.87$\pm$0.13  \\
\multicolumn{1}{l}{\textbf{SIMI}\textsubscript{q+a+e}}& 18.89$\pm$2.35 & 6.17$\pm$2.03 & 13.79$\pm$2.20 & 1.54$\pm$0.64  \\
\multicolumn{1}{l}{\textbf{MHOP}\textsubscript{q+a+e}}& 21.35$\pm$2.57 & 6.79$\pm$2.31 & 14.77$\pm$1.98 & 1.83$\pm$0.57   \\
\multicolumn{1}{l}{\textbf{MEAN}\textsubscript{q+a+e}}& 22.57$\pm$2.50 & 7.13$\pm$1.78 & 16.04$\pm$1.65 & 2.25$\pm$0.93    \\
\hline
\multicolumn{1}{l}{\textbf{EMIN}} & \textbf{32.24}$\pm$3.21 & \textbf{13.23}$\pm$2.98 & \textbf{23.19}$\pm$2.79 & \textbf{10.48}$\pm$3.22 \\

\bottomrule
\end{tabular*}
\end{center}
\captionsetup{justification=centering}
\caption{Experimental results in terms of ROUGE (R) and BLEU scores (\%) on the MQAE-diag dataset. ``q'', ``a'', and ``e'' represents questions, answers and evidence, respectively. 
}
\label{table:diag_qae_test}
\end{table*}

\subsection{Experimental Set-up}

\noindent \textbf{Model Settings:} 
The trained QA system is the BioLinkBERT~\cite{yasunaga2022linkbert}.
We employ the system on the four-options multiple question-answering tasks and it achieves an accuracy of $60.18 \%$ on the MQAE-diag dataset and $55.85\%$ on the MQAE dataset.
The maximum length of questions and evidence spans is $944$ and the explanations' maximum length is $128$.
The size of the hidden representations of the model is $768$. The number of selected evidence paragraphs $k$ is set to 10 (we evaluate the influence of the value of this hyperparameter).

\noindent \textbf{Training and Inference Details:} 
\textbf{1)} For the training, all baseline models and our framework are trained with fp16.
All of the models utilize the AdamW optimizer~\cite{loshchilov2017decoupled} with a learning scheduler initialized at $2e^{-5}$ and linearly decreased to 0.
We apply the early stopping strategy on the baseline models by monitoring the loss on the validation set while the patience number is set to $4$. 
The KL divergence threshold $\epsilon$ for our approach in the training stage is~$0.01$. 
\textbf{2)} 
During the inference stage, we use beam search  to generate text and set the width of the beam to~$5$.
$\epsilon$ in the inference stage is also~$0.01$.

\begin{table*}[htbp]
\setlength\tabcolsep{3pt} 
\begin{center}
\begin{tabular*}{0.9\linewidth}{@{\extracolsep{\fill}} l r|r|r|r }
\toprule
Models & \multicolumn{1}{l}{R1} & \multicolumn{1}{l}{R2} & \multicolumn{1}{l}{RL} & \multicolumn{1}{l}{BLEU-4} \\
\midrule
\multicolumn{1}{l}{\textbf{T5}\textsubscript{q+a}}&  14.48$\pm$1.43 & 5.94$\pm$1.01 & 13.33$\pm$1.34 & 0.30$\pm$0.26 \\
\multicolumn{1}{l}{\textbf{BART}\textsubscript{q+a}} &  15.61$\pm$1.44 & 6.49$\pm$1.47 & 13.47$\pm$1.38 & 0.68$\pm$0.24 \\
\multicolumn{1}{l}{\textbf{SIMI}\textsubscript{q+a+e}}&  18.18$\pm$2.03 & 7.87$\pm$1.67 & 15.21$\pm$1.94 & 0.82$\pm$0.37 \\
\multicolumn{1}{l}{\textbf{MHOP}\textsubscript{q+a+e}}&  20.03$\pm$2.15 & 8.07$\pm$1.98 & 16.78$\pm$1.97 & 1.03$\pm$0.56  \\
\multicolumn{1}{l}{\textbf{MEAN}\textsubscript{q+a+e}}&  21.17$\pm$2.03 & 9.57$\pm$2.12 & 17.71$\pm$1.84 & 1.35$\pm$0.43  \\
\hline
\multicolumn{1}{l}{\textbf{EMIN}} &   \textbf{28.03}$\pm$2.81 & \textbf{12.77}$\pm$3.06 & \textbf{21.08}$\pm$2.34 & \textbf{9.17}$\pm$2.89 \\

\bottomrule
\end{tabular*}
\end{center}
\captionsetup{justification=centering}
\caption{Experimental results in terms of ROUGE (R) and BLEU scores (\%) on the MQAE datasets. ``q'', ``a'', and ``e'' represents questions, answers and evidence, respectively. 
}
\label{table:mqae_test}
\end{table*}

\subsection{Baseline Models}

Due to the maximum length of the BART encoder and other language model based encoders, we cannot use the concatenation of the question, answer and and all selected evidences as the input. Hence the simple baselines below do not use retrieved evidence paragraphs. Their results are mentioned as a kind of underbound of the performance. The generated explanation here only relies on the pretrained language model.


\noindent \textbf{T5~(q+a)}~\cite{raffel2020exploring} passes the concatenation of questions and answers to the T5 language model to generate explanations.


\noindent \textbf{BART~(q+a)}~\cite{lewis2019bart} simply concatenates questions and answers as input, which is fed into the BART model to generate explanations.

\noindent The following models retrieve paragraphs in the same way as done in our model (see \ref{sec:evidence_select}), and the evidence paragraphs are individually encoded with the BART encoder (see \ref{sec:nn-enc}).

\noindent \textbf{SIMI~(q+a+e)}
\label{sec:simi}
removes EMIN's E-step and computes the evidence weights $\mathbf{z_i^\prime}$ with the dot product similarity between the encoder outputs of question-answer representations and evidence features normalized by a softmax function.
SIMI leverages the EMIN's decoder to incorporate evidence. 

\noindent \textbf{MHOP~(q+a+e)} 
\label{sec:mhop}
Following the literature on Key-Value Memory Networks~\cite{miller2016key}, MHOP encodes the evidence paragraphs with the BART encoder and attends in multiple hops to this encoded "memory". 

\noindent \textbf{MEAN~(q+a+e)} 
\label{sec:mean}
discards the E-step in EMIN, and uses uniform evidence weights to fuse the evidence features, so evidence weights are not learned with EM inference. 
The dual encoder-decoder is the same as for EMIN.


\subsection{Results and Discussion}

To compare with baseline models, we report the results of models on the ROUGE
~(recall oriented metric and coded further as R) and BLEU-4
scores~(precision oriented metric) that compare the ground-truth explanation with the generated explanation. The numbers in the metric names refer to the size of n-gram of words considered. ROUGE-L (RL) takes into account the length of the longest common subsequence of words.
Table~\ref{table:diag_qae_test} and Table~\ref{table:mqae_test} reports test set performance for all models on both datasets.

\noindent \textbf{MQAE-diag.} Our model outperforms the baseline models by large margins across all evaluation metrics.
Compared to the baseline models that use encoded evidence paragraphs to generate explanations (i.e., {SIMI}, {MHOP}, and {MEAN}), EMIN improves the BLEU score by $11.5$, $9.99$, and $9.75$ percentage points, respectively.
{MEAN} is the best evidence-enhanced model, and EMIN outperforms it by $9.87$, $6.61$, $7.28$, and $9.73$ percentage points on the R1, R2, RL and BLEU scores, respectively.
EMIN beats the baselines that do not consider evidence paragraphs by even larger margins. 



\noindent \textbf{MQAE.} 
On the MQAE dataset, EMIN improves all evaluation scores, especially the BLEU score.
Our approach outperforms the best no-evidence model, i.e., {BART (q+a)}, by $14.09$, $7.77$, $10.23$, and $11.35$ percentage points on the R1, R2, RL and BLEU scores, respectively.
In particular, our framework improve the R1 scores by $9.27$, $8.00$, and $6.45$ percentage points compared with the evidence-enhanced models~({SIMI}, {MHOP}, and {MEAN}).
The results show that EMIN yields large gains over all competitors and the evidence based approaches are all much better than the non evidence based ones. 

\noindent \textbf{Discussion:}  we compare the results presented in Table~\ref{table:diag_qae_test} and Table~\ref{table:mqae_test} to emphasize the significance of our proposed EM (Expectation-Maximization) inference method. 
Unlike the MEAN model, which employs uniform weights throughout, EMIN iteratively updates evidence weights. 
Essentially, MEAN can be seen as the starting point of EMIN, where there is no distinction in assigning importance to each evidence paragraph when generating explanations.
While both the MHOP and EMIN models iteratively adjust attention weights on the tokens within individual evidence paragraphs within the dual encoder-decoder architecture, it is worth noting that MHOP does not capture the importance of larger evidence paragraphs. 
Here, EM inference offers an elegant solution for reasoning at the paragraph level.
To demonstrate the efficiency of EM inference, we conducted a comparison between EM inference and a model using full attention across concatenated evidence paragraphs.
Assume that we have $m$ paragraphs and each paragraph has $n$ words, the time complexity of the EM inference and the full attention one are $\mathcal{O} (m^2 \cdot n)$ and $\mathcal{O} (m^2 \cdot n^2)$, respectively.
Consequently, EM inference proves to be more efficient than the full attention model.

\subsection{Case Study}
Table~\ref{table:error_case} presents a case study on explanation generation in the context of a trained QA system making a wrong prediction.
From the table, we can observe that the ground-truth explanation and EMIN explanation (based on the ground-truth answer) are all indicate ``metformin increase insulin output''.
Even though the trained QA systems make a incorrect prediction, i.e., ``acarbose'', the EMIN explanation (based on the incorrect answer) still suggests the ``metformin'' is the correct drug for the type 2 diabetes patients. 

\begin{table*}[htbp]
\centering
	
\begin{tabularx}{\textwidth}{|X|}
\toprule
\textbf{Question:} \small a 52-year-old woman begins pharmacotherapy after being diagnosed with type 2 diabetes mellitus. four weekslater, her hepatic glucose output is decreased, and target tissue glucose uptake and utilization are increased. whichof the following drugs was most likely prescribed for this patient? \\
\midrule
\textbf{GT Explanation:} \small metformin is awesome. it decreases hepatic glucose production, decreases intestinal absorption of glucose, and improves insulin sensitivity by increasing peripheral glucose uptake and utilization. \\
\midrule
\textbf{EMIN Explanation~(GT):} \small metformin (a type 2 diabetes) is a ketoacidotic with a tendency to produce hepatic dysfunction and increase insulin output. It is a polypeptide hormone, so it would be unlikely for the patient to be an insulinoma type 2.As a side effect, it blocks insulin production and decreased glucose output.\\
\midrule
\textbf{EMIN Explanations~(WA):} \small Acarbose is a glycoprotein that binds to glucose receptors in the liver, causing it to decrease hepatic glucose output and increase protein synthesis. Inhibitors to glycerol/sulfamethoxazole (a type 2 diabetes antagonist), clofibrate, and \textit{mefenamic acid are the most useful drugs for Type 2 diabetes patients}.\\
\bottomrule
	\end{tabularx}
	\caption{As an example from the MCQAE dataset
 The correct answer for this question is `metformin', but the trained QA system predicts `acarbose' as the answer. In this context, `GT' refers to the explanation based on the ground-truth answer, and `WA' represents the explanation based on the wrong answer.
 The Italian words in EMIN explanation explanations~(WA) indicated that the ``metformin'' is the correct drug for the type 2 diabetes patients. 
 }
	\label{table:error_case}
\end{table*}%

\section{Related Work}
\label{sec:related}

In recent years, significant advancements in medical question-answering systems have been driven by the introduction of novel models and datasets~\cite{abacha2015means,veisi2020persian,pal2022medmcqa}.
However, within the domain of medical inquiry, the role of explanations has become increasingly critical. 
Approaches such as segmenting highlighted textual spans~\cite{hu2021explainable} and knowledge-base paths~\cite{zhang_exp} have limitations when they come to providing explanation which requires substantial mental effort, particularly because out-of-context spans may be difficult to interpret. 
An alternative approach would be to generate a natural language explanation.

Large Language Models (LLMs) have achieved remarkable performance on medical question-answering datasets, emphasizing the significance of accurate predictions.
However, user acceptance, particularly among clinicians, is contingent upon the model's ability to provide transparent explanations of its decision-making process.
To address this need, Chain-of-thought~(CoT)-based techniques \cite{wei2022chain,zhou2022least} are proposed. 
In the medical domain, applications deployed in hospitals must prioritize data privacy, necessitating local server deployment.
However, the sheer size of LLMs precludes their local deployment in such contexts.

\section{Conclusions}
\label{sec:conclude}

We propose an approach to collectively reason over all textual evidence while harnessing the potent attention mechanisms of a transformer architecture. 
Our method employs an Expectation-Maximization~(EM) algorithm to deduce the weights of evidence paragraphs as latent variables, enhancing the model parameters for explanation generation. 
This gives rise to the EMIN framework, which notably outperforms existing models, such as those utilizing multi-hop reasoning, in terms of generating explanations.
The explicit evidence weights we derive unveil which pieces of evidence bolster the generated explanations. 
These weights complement the attention weights assigned to tokens within the evidence paragraphs, all in a scalable manner. 
This ensures that users gain comprehensive insights into the significance of the evidence within the generated explanation.

\bibliographystyle{splncs04}
\bibliography{main}

\end{document}